\documentclass{article}

\usepackage[nonatbib,preprint]{neurips_2023}

\usepackage{array}
\usepackage{subcaption}

\newcommand{\tablestyle}[2]{\setlength{\tabcolsep}{#1}\renewcommand{\arraystretch}{#2}\centering\footnotesize}

\newlength\savewidth\newcommand\shline{\noalign{\global\savewidth\arrayrulewidth
  \global\arrayrulewidth 1pt}\hline\noalign{\global\arrayrulewidth\savewidth}}

\usepackage{wrapfig}

\usepackage[utf8]{inputenc} % allow utf-8 input
\usepackage[T1]{fontenc}    % use 8-bit T1 fonts
\usepackage{hyperref}       % hyperlinks
\usepackage{url}            % simple URL typesetting
\usepackage{booktabs}       % professional-quality tables
\usepackage{amsfonts}       % blackboard math symbols
\usepackage{nicefrac}       % compact symbols for 1/2, etc.
\usepackage{microtype}      % microtypography
\usepackage{xcolor}         % colors
\usepackage{times}
\usepackage{epsfig}
\usepackage{graphicx}
\usepackage{amsmath}
\usepackage{amssymb}
\usepackage{bbm}
\usepackage{multirow}
\usepackage{color}
\usepackage{colortbl}
\usepackage{xcolor}
\usepackage{multirow}
\usepackage{makecell}
\usepackage{tikz}
\usepackage{bbding}
\usepackage[capitalize]{cleveref}
\usepackage{expl3}
\ExplSyntaxOn
\newcommand\latinabbrev[1]{
	\peek_meaning:NTF . {% Same as \@ifnextchar
		#1\@}%
	{ \peek_catcode:NTF a {% Check whether next char has same catcode as \'a, i.e., is a letter
			#1.\@ }%
		{#1.\@}}}
\ExplSyntaxOff

\def\eg{\latinabbrev{e.g}}
\def\etal{\latinabbrev{et al}}

\def\ie{\latinabbrev{i.e}}

\def\wrt{\latinabbrev{w.r.t}}

\DeclareMathOperator*{\argmax}{arg\,max}

\definecolor{applegreen}{rgb}{0.0, 0.5, 0.0}

\definecolor{aliceblue}{rgb}{0.94, 0.97, 1.0}

\title{PaintSeg: Training-free Segmentation via Painting}

\author{%
Xiang Li \\
CMU
  \And
  Chung-Ching Lin \\
  Microsoft
  \And
  Yinpeng Chen \\
  Microsoft
  \AND
  Zicheng Liu \\
  Microsoft
  \And
  Jinglu Wang \\
  Microsoft
  \And
  Bhiksha Raj \\
  CMU \& MBZUAI
}

\begin{document}

\maketitle
\def\tabdesignchoice#1{
\begin{table*}[#1]
	\centering
	\subfloat[
	\textbf{Cluster center}.
	\label{tab:cluster center}
	]{
		\centering
		\begin{minipage}{0.2\linewidth}{\begin{center}
    \tablestyle{1.5pt}{1.2}
    \begin{tabular}{l|ccc}
        % \shline
        K & 2 & 3 & 4 \\ [.1em]
        \shline
    IoU & \bf80.6 & 72.3 & 61.5 \\
        % \hline
    \end{tabular}
		\end{center}}\end{minipage}
	}
	\subfloat[
	\textbf{Step number.}.
	\label{tab:step number}
	]{
		\begin{minipage}{0.26\linewidth}{\begin{center}
    \tablestyle{1.5pt}{1.2}
    \begin{tabular}{l|ccccc}
        % \shline
        $T$ & 3 & 4 & 5 & 6 \\ [.1em]
        \shline
        IoU &  78.4 & 79.1 & \bf80.6 & 80.5 \\
        % \hline
    \end{tabular}
\end{center}}\end{minipage}
}
	\subfloat[
	\textbf{Iter. for painting}.
	\label{tab:iter painting}
	]{
		\begin{minipage}{0.2\linewidth}{\begin{center}
					\tablestyle{1.5pt}{1.2}
					\begin{tabular}{l|ccc}
                        % \shline
                        Iter & 10 & 30 & 50 \\ [.1em]
                        \shline
					IoU & 77.3 & 78.7 & \bf80.6 \\
                        % \hline
                    \end{tabular}
		\end{center}}\end{minipage}
	}
    \subfloat[
	\textbf{Box size for contrasting}.
	\label{tab:box size}
	]{
		\begin{minipage}{0.3\linewidth}{\begin{center}
					\tablestyle{1.5pt}{1.2}
					\begin{tabular}{l|cccc}
                        % \shline
                        Rate & 0.9 & 1.0 & 1.1 & 1.2 \\ [.1em]
                        \shline
						IoU & 69.2 & 72.4 & \bf80.6 & 80.0 \\
                        % \hline
                    \end{tabular}
		\end{center}}\end{minipage}
	}
 % \vspace{-5pt}
	\caption{\textbf{Design choices for AMCP.} We report the performance with the coarse-mask prompt on ECSSD. (a) We ablate the cluster center when contrasting. (b) We ablate the step number for AMCP. (c) We ablate the diffusion iteration for generative painting. (d) We ablate on the cropped box size when contrasting. The rate denotes the proportion of cropped mask and the box of the current stage object mask.}
	\label{tab:design choices}
\end{table*}}

\def\tabeffectiverobust#1{
\begin{table*}[#1]
\begin{minipage}[t]{\textwidth}
\begin{minipage}[t]{0.38\textwidth}
\makeatletter\def\@captype{table}
\centering
\scalebox{0.8}{
\begin{tabular}{cccc} 
\toprule
I-step & O-step & AC & IoU\\
\midrule 
\Checkmark & & & 78.4 \\
& \Checkmark & & 77.9 \\
\Checkmark & \Checkmark & & 79.5 \\
\Checkmark & \Checkmark & \Checkmark & 80.6 \\
\bottomrule
\end{tabular}
}
\caption{\textbf{Module effectiveness in AMCP}. AC: adversarial constraint for mask updating.}
\label{tab:effectiveness}
\end{minipage}
\hspace{0.01\textwidth}
\begin{minipage}[t]{0.6\textwidth}
\makeatletter\def\@captype{table}
\centering
\scalebox{0.89}{
\begin{tabular}{lccccccccccccc}
    \toprule
         Prompt & 0\% Noise & 15\% Noise & 30\% Noise \\
        \midrule
         Point & 60.8 & 60.4 & 58.9 \\
         Scribble & 64.3 & 63.7 & 62.7 \\
         Box & 71.0 & 70.7 & 70.1 \\
         Coarse Mask & 80.6 & 80.0 & 79.3 \\
        \bottomrule
        \end{tabular}}
        \caption{\textbf{Prompt robustness.} We add random noise to the prompt to evaluate the robustness of AMCP. The noise scale is determined by half the length of the object box diagonal. 
    }
    \vspace{-5pt}
    \label{tab:prompt sensitivity}
\end{minipage}
\end{minipage}
\vspace{-5pt}
\end{table*}}

\def\tabboxprompt#1{
\begin{table*}[#1]
    \centering
    \resizebox{0.8\textwidth}{!}{
    \begin{tabular}{lcccccccccc}
    \toprule
    Method & Training & Supervision & PASCAL VOC &  MVal \\
    \multicolumn{5}{c}{--- \textit{Compared to methods with training} ---} \\
        Mask-RCNN \cite{he2017mask}$\,_{\mathrm{ICCV}17}$ & \checkmark & \checkmark & \bf73.2 & \bf79.4 \\
        CutLER \cite{wang2023cut}$\,_{\mathrm{CVPR}23}$ & \checkmark & & 63.5 & 74.8 \\
        \rowcolor{aliceblue!80}\bf{PaintSeg} & & & 59.7 & 69.6\\
    \multicolumn{5}{c}{--- \textit{Compared to methods without training} ---} \\
        TokenCut \cite{wang2022tokencut}$\,_{\mathrm{CVPR}22}$ & & & 30.2 & 34.7 \\
        \rowcolor{aliceblue!80}\bf{PaintSeg} & & & \bf59.7 & \bf69.6 \\
        \bottomrule
        \end{tabular}}
        \caption{\textbf{Qantitative comparison of box-prompted segmentation on PASCAL VOC and COCO MVal.} 
    }
    \vspace{-10pt}
    \label{tab:box prompt}
\end{table*}}

\def\tabpointprompt#1{
\begin{table*}[#1]
    \centering
    \resizebox{0.8\textwidth}{!}{
    \begin{tabular}{lcccccccccc}
    \toprule
    Method & Training & Supervision & GrabCut & Berkeley & DAVIS \\
            \multicolumn{6}{c}{--- \textit{Compared to methods with training} ---} \\
        DIOS$ \cite{xu2016deep}\,_{\mathrm{CVPR}16}$ & \checkmark & \checkmark & 64.0 & 66.0 & 57.8 \\
        RITM \cite{sofiiuk2022reviving}$\,_{\mathrm{ICIP}22}$ & \checkmark & \checkmark & 81.0 & \bf77.7 & 66.0 \\
        MIS \cite{li2023multi}$\,_{\mathrm{arXiv}23}$ & \checkmark & & 76.2 & 63.2 & 53.3 \\
        \rowcolor{aliceblue!80}\bf PaintSeg & & & \bf84.4 & 70.0 & \bf69.4 \\
     \multicolumn{6}{c}{--- \textit{Compared to methods without training} ---} \\
        Random Walk \cite{grady2006random}$\,_{\mathrm{TPAMI}06}$ & & & 25.7 & 26.2 & <20\\
        GrowCut \cite{vezhnevets2005growcut}$\,_{\mathrm{GraphiCon}05}$ & & & 26.7 & 26.2 & - \\
        GraphCut \cite{boykov2001interactive}$\,_{\mathrm{ICCV}01}$ & & & 41.8 & 33.9 & <20 \\
        \rowcolor{aliceblue!80}\bf PaintSeg & & & \bf84.4 & \bf70.0 & \bf69.4\\
        \bottomrule
        \end{tabular}}
        \caption{\textbf{Qantitative comparison of point-prompted segmentation on GrabCut, Berkeley, and DAVIS.} The point prompt is given as the centroid of each object.
    }
    \vspace{-5pt}
    \label{tab:point prompt}
\end{table*}
}

\def\tabmaskprompt#1{
\begin{table*}[#1]
    \centering
    \resizebox{0.8\textwidth}{!}{
    \begin{tabular}{lp{2.4cm}<{\centering}p{2.4cm}<{\centering}p{2.4cm}<{\centering}}
    \toprule
         Method & Training & DUTS-TE \cite{wang2017} & ECSSD\cite{shi2016ecssd}\\
        \midrule
        \multicolumn{4}{c}{--- \textit{Compared to methods with training} ---} \\
         SelfMask \cite{shin2022selfmask}$\,_{\mathrm{CVPRW}22}$ & \checkmark & {62.6} & {78.1} \\
         SelfMask \cite{shin2022selfmask}$\,_{\mathrm{CVPRW}22}$ + BS \cite{barron2016fast} & \checkmark & {66.0} & \bf{81.8} \\
         FOUND \cite{simeoni2022unsupervised}$\,_{\mathrm{CVPR}23}$ & \checkmark & {63.7} & {79.3} \\ 
         FOUND \cite{simeoni2022unsupervised}$\,_{\mathrm{CVPR}23}$ + BS \cite{barron2016fast} & \checkmark & \underline{66.3} & {80.5} \\ 
    \rowcolor{aliceblue!80}\bf PaintSeg & & \bf{67.0} & \underline{80.6} \\
         \multicolumn{4}{c}{--- \textit{Compared to methods without training} ---} \\
         % HS \cite{yan2013hs}$\,_{\mathrm{CVPR}13}$ &  & 36.9 & 50.8 \\
         % wCtr \cite{zhu2014wctr}$\,_{\mathrm{CVPR}14}$ & & 39.2 & 51.7 \\
         % WSC \cite{li2015wsc}$\,_{\mathrm{CVPR}15}$ & & 38.4 & 49.8 \\
         % DeepUSPS \cite{nguyen2019deepusps}$\,_{\mathrm{NeurIPS}19}$ & & 30.5 & 44.0\\
         % BigBiGAN \cite{voynov2021biggan}$\,_{\mathrm{ICML}21}$  & & 49.8 & 67.2 \\
         % E-BigBiGAN \cite{voynov2021biggan}$\,_{\mathrm{ICML}21}$  & & 51.1 & 68.4 \\
         Melas-Kyriazi et al. \cite{melas2021}$\,_{\mathrm{ICLR}22}$\quad\quad & & 52.8 & 71.3 \\
         LOST \cite{simeoni2021lost}$\,_{\mathrm{BMVC}21}$ & & 51.8 & 65.4 \\
         LOST \cite{simeoni2021lost}$\,_{\mathrm{BMVC}21}$ + BS \cite{barron2016fast} & & 57.2 & 72.3 \\
         DSS \cite{melas2022deepsectralmethod}$\,_{\mathrm{CVPR}22}$ & & 51.4 & 73.3 \\
         TokenCut \cite{wang2022tokencut}$\,_{\mathrm{CVPR}22}$ & & 57.6 & 71.2 \\
         TokenCut \cite{wang2022tokencut}$\,_{\mathrm{CVPR}22}$ + BS \cite{barron2016fast} & & \underline{62.4} & \underline{77.2} \\
         SelfMask$\dagger$ \cite{shin2022selfmask}$\,_{\mathrm{CVPRW}22}$ & & 46.6 & 64.6 \\
         FOUND$\dagger$ \cite{simeoni2022unsupervised}$\,_{\mathrm{CVPR}23}$ & & - & 71.7 \\
         \rowcolor{aliceblue!80}\bf PaintSeg & & \bf{67.0} & \bf{80.6} \\

         \bottomrule
        \end{tabular}}
        \caption{\textbf{Qantitative results of coarse mask-prompted segmentation on DUTS-TE and ECSSD.} PaintSeg utilizes the coarse mask generated by unsupervised TokenCut \cite{wang2022tokencut} as prompt. BS denotes the application of the post-processing bilateral solver on the generated masks and the column `Learning' specifies which methods have a training step. The best result per section is highlighted in \textbf{bold}. The second best result for each section is underlined.  $\dagger$ indicates the first-stage pseudo mask obtained without training.
    }
    \vspace{-5pt}
    \label{tab:saliency-detection}
\end{table*}}

\begin{abstract}
The paper introduces PaintSeg, a new unsupervised method for segmenting objects without any training. We propose an adversarial masked contrastive painting (AMCP) process, which creates a contrast between the original image and a painted image in which a masked area is painted using off-the-shelf generative models. During the painting process, inpainting and outpainting are alternated, with the former masking the foreground and filling in the background, and the latter masking the background while recovering the missing part of the foreground object. Inpainting and outpainting, also referred to as I-step and O-step, allow our method to gradually advance the target segmentation mask toward the ground truth without supervision or training. PaintSeg can be configured to work with a variety of prompts, e.g. coarse masks, boxes, scribbles, and points. Our experimental results demonstrate that PaintSeg outperforms existing approaches in coarse mask-prompt, box-prompt, and point-prompt segmentation tasks, providing a training-free solution suitable for unsupervised segmentation.
\end{abstract}

\section{Introduction}

With deep learning advancements, significant progress has been made in the field of image generation and segmentation in recent years. A particular generative model, the denoising diffusion probabilistic model (DDPM), has demonstrated outstanding performance in a variety of generative tasks, such as image inpainting \cite {rombach2021highresolution,dong2022incremental} and text-to-image synthesis \cite{galatolo2021generating, gal2022stylegan, zhou2021lafite}. Similar developments have occurred in the field of object segmentation, such as the strong zero-shot capability and excellent segmentation quality demonstrated by SAM \cite{kirillov2023segment}. 
% However, all these segmentation models require elaborate training.

Image generation and segmentation can be mutually beneficial. Segmentation has been shown to be a critical technique in improving the realism and stability of generative models by providing pixel-level guidance during the synthesis process \cite{zhang2023adding,isola2017image}. Interesting to note is the fact that the relationship between segmentation and generative models does not appear to be solely one-sided. Generative models learning to ``paint" objects actually know where the painted object is. The emergence of unsupervised image segmentation methods utilizing generative adversarial networks (GANs) has produced a line of methods that can segment objects in images \cite{benny2020onegan,chen2019unsupervised,bielski2019emergence} using generative models. These methods work on the assumption that object appearance and location can be perturbed without compromising scene realism. By using the GAN architecture to discriminate between perturbed and real images, these methods can achieve effective object segmentation. Moreover, a follow-up work \cite{voynov2021biggan} develops an approach to leverage pre-trained GAN by identifying ``segmenting" direction in the latent space to discriminate object shapes. 

In this paper, we present PaintSeg, an approach for unsupervised image segmentation that leverages off-the-shelf generative models. Unlike previous methods \cite{voynov2021biggan,benny2020onegan} that require training on top of these models, PaintSeg introduces a novel, training-free segmentation approach that relies on an adversarial masked contrastive painting (AMCP) process. The AMCP process creates a contrast between the original image and a painted image by alternating between inpainting and outpainting, with the former filling in the background and masking the foreground, and the latter retrieving the missing part of the object while masking the background and a portion of the foreground. 

\begin{figure}[t]
    \centering    
    \includegraphics[width=\linewidth]{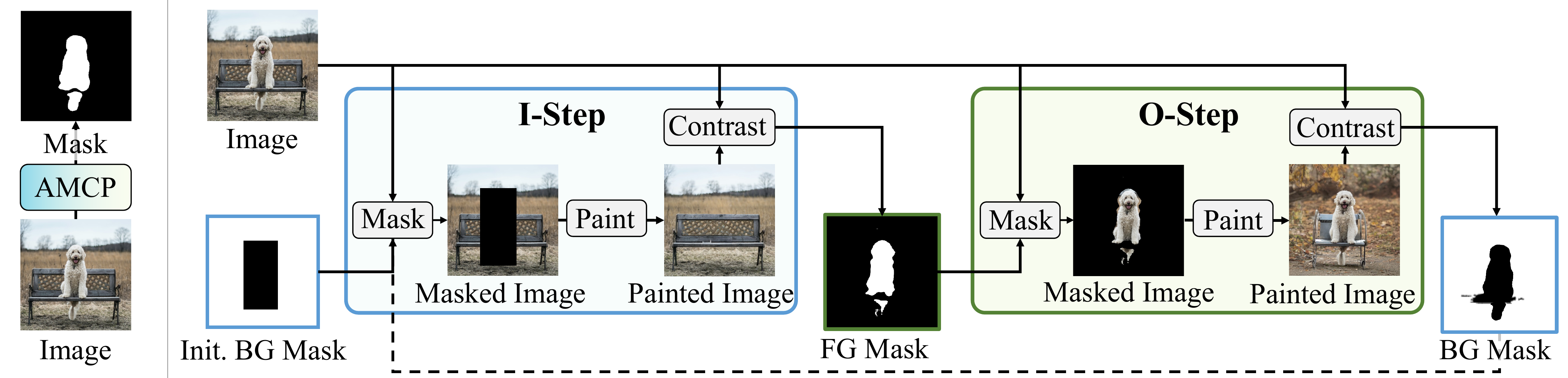}
    \caption{\textbf{Illustration of adversarial masked contrastive painting (AMCP).}  Given an input image and an initial mask, AMCP leverages alternating I-step and O-step to gradually refine the segmentation mask until it converges to the ground truth. Both steps share the same mask, paint, and contrast operations. The updated mask in each step is achieved by binarizing the contrastive difference between the original and painted images. 
    }
    \label{fig:teaser}
\end{figure}

Both steps, as shown in \cref{fig:teaser}, share the same operations while taking input from background and foreground masks, correspondingly. In the I-step, the object region is removed from the painted image, creating a significant contrast with the original image. Conversely, in the O-step, the background region exhibits a remarkable difference between the original and painted image. The foreground or background mask can be obtained by binarizing the contrastive difference in each step.

Although either I-step or O-step is capable of discriminating objects, the single-step method is less robust. The I-step involves segmenting objects based on background consistency without taking into account object information. As a result, the segmentation may be imperfect if the object part resembles the background. Similarly, in the O-step, only the object shape prior is utilized, resulting in a lack of background knowledge. This problem is addressed by introducing adversarial mask updating, in which I-steps and O-steps are alternated. During I-step, we only shrink the object mask to cut off background false positives, while during O-step, we expand it to link up foreground false negatives. Thereby, even if errors occur during the iteration of AMCP, they will be corrected in the next step without degradation. With the adversarial mask updating, the target mask can be gradually advanced to the ground truth.

With the robustness of AMCP, PaintSeg can deal with inaccurate initial masks and adapt to various visual prompts, such as coarse masks, bounding boxes, scribbles, and points. Compared to the recently published successes in image object segmentation study, our main contributions are as follows:
\begin{itemize}
    \item 
    % We propose PaintSeg, a training-free approach for image object segmentation adapting to heterogeneous visual prompts, by bridging the generative model with segmentation.
    % We propose PaintSeg, a training-free approach for image object segmentation that adapts to heterogeneous visual cues and bridges generative models with segmentation.
    We propose PaintSeg, a training-free approach to segmenting image objects based on heterogeneous visual cues. The method provides a direct bridge between generative models and segmentation.
    \item We introduce adversarial masked contrastive painting (AMCP), consisting of alternating I-step and O-step, to robustly segment objects. 
    \item We conduct extensive experiments for analysis and comparisons on seven different image segmentation datasets, the results of which show the superiority and generalization ability of our methods.
\end{itemize}

% Reducing the data reliance for image segmentation has been extensively researched in the field of computer vision to avoid the time-consuming and costly human annotation. The recent studies in self-supervised learning, such as DINO \cite{caron2021dino,oquab2023dinov2}, observe a strong object-level pixel grouping though their training objective is fully label-free, which motivates several research works \cite{wang2022tokencut,wang2023cut,simeoni2023found} to segment objects by leveraging self-supervised image representations. 
% Before that, another line of unsupervised image segmentation methods utilize generative adversarial networks (GAN) \cite{benny2020onegan,chen2019unsupervised,bielski2019emergence} to segment objects with the assumption that object appearance/location can be perturbed without losing scene realism. Thereby, segmentation can be achieved by discriminating between perturbed and real images using the GAN architecture.

% In this paper, we focus on unsupervised image segmentation task, and aim to explore the usage of implicit object shape encoded in generative painting models \cite{rombach2021highresolution}. Object segmentation aims to group semantically closed 

% 强调bridge了 generative model，shape prior，spatial smoothness

% 强调不是简单的利用shape prior，我们用的了xxx，adversarial

% 第二段少写点，用了training
\section{Related Works}
\subsection{Unsupervised Image Segmentation} 
Unsupervised methods for image segmentation are extensively investigated with the advancements in self-supervised. DINO~\cite{caron2021emerging} provides a self-supervised approach to explicitly bring out underlying semantic segmentation of images using a Vision Transformer (ViT)~\cite{dosovitskiy2020image}.
Based on DINO, LOST~\cite{simeoni2021lost}, Deep Spectral Methods \cite{melas2022deep} and TokenCut~\cite{wang2022tokencut} leverage self-supervised ViT features and propose to segment objects using NCut \cite{shi2000normalized}. Subsequently, \cite{simeoni2022unsupervised,shin2022selfmask} introduce a second-stage training approach to further improve the segmentation quality. Found \cite{simeoni2022unsupervised} incorporates background similarity as an additional refinement factor, while SelfMask \cite{shin2022selfmask} utilizes an ensemble of features \cite{caron2020unsupervised,caron2021emerging,chen2020improved} to enhance image representation. CutLER \cite{wang2023cut} enables multiple objects discovery capability by iteratively cutting objects with NCut and introduces a more powerful second-stage training.
FreeSOLO~\cite{wang2022freesolo} generates coarse masks with correlation maps that are then ranked and filtered by a ``maskness" score.
Another line of unsupervised methods learns to generate a realistic image by combining a foreground, a background and a mask \cite{bielski2019emergence,yang2019unsupervised,yang2017lr,van2020investigating,kwak2016generating,eslami2016attend,he2022ganseg} and then the object segmentor can be obtained as a byproduct.

\subsection{Prompt-guided Segmentation}
Prompt-guided segmentation aims to segment objects assigned by prompts, \eg, mask, box, scribble and point. Semi-supervised video object segmentation (VOS) \cite{athar2023tarvis,xie2021efficient,liang2020video}, aiming at segmenting object masks across frames given the first frame mask, is a typical mask-prompt task. The mainstream of VOS methods ~\cite{yang2020cfbi,yang2021associating} constructs pixel-level correspondence and propagates masks by exploring matches among adjacent frames. Interactive segmentation (IS) \cite{zhou2023interactive,lin2022focuscut,sofiiuk2022reviving,hao2021edgeflow} is another line of prompt-guided segmentation. IS permits users to leverage scribbles and points to assign target objects and segment them. In addition, an interactive correction is also featured by IS which introduces additional prompts to correct misclassified regions. MIS \cite{li2023multi} is a recent work tackling unsupervised IS and proposes a multi-granularity region proposal generation to refine the mask. SAM \cite{kirillov2023segment} is a recently introduced zero-shot method for prompt-based segmentation which introduces a large-scale dataset and a strategy to mitigate the ambiguity of prompt. Beyond visual prompts, objects can also be referred by natural language or acoustic prompts. Referring image segmentation (RIS) \cite{hu2016segmentation, yu2018mattnet} and referring video object segmentation (R-VOS) \cite{chen2022multi,zhao2022modeling,ding2022language,li2022r} aims to segment objects in image/video referred by linguistic expressions. Audiovisual segmentation \cite{zhou2022avs} aims to segment sound sources in the given audiovisual clip.

\subsection{Conditional Image Generation}

Conditional image generation refers to the process of generating images based on specific conditions or constraints. In most instances, the condition can be based on class labels, partial images, semantic masks, etc. Cascaded Diffusion Models \cite{ho2022cascaded} uses ImageNet class labels as a condition to generate high-resolution images with a two-stage pipeline of multiple diffusion models. \cite{sehwag2022generating} guides diffusion models to produce novel images from low-density regions of the data manifold. Apart from these, CLIP \cite{radford2021clip} has been widely used in guiding image generation in GANs with text prompts \cite{galatolo2021generating, gal2022stylegan, zhou2021lafite}. For diffusion models, Semantic Diffusion Guidance \cite{liu2023more} investigates a unified framework for diffusion-based image generation with language, image, or multi-modal conditions. Dhariwal et al. \cite{dhariwal2021diffusion} apply an ablated diffusion model to use the gradients of a classifier to guide the diffusion with a trade-off between diversity and fidelity. Additionally, Ho \etal \cite{ho2022classifier} introduce classifier-free guidance in conditional diffusion models by mixing the score estimates of a conditional diffusion model and a jointly trained unconditional diffusion model.

\section{Problem Definition}
We tackle the unsupervised prompt-guided image object segmentation task, which aims to predict the object mask $M\in\{0,1\}^{1\times H\times W}$ in an image $I\in\mathbb{R}^{3\times H\times W}$ given a visual prompt $P\in\{0,1\}^{1\times H\times W}$. The visual prompt can have a format of a point, a scribble, a bounding box or a coarse mask of the target object $P\in\{P_{point},P_{scrib},P_{box},P_{mask}\}$. Following the convention, we assume the ground-truth object mask $M$ must have an overlap with the visual prompt $P\cap M\neq\emptyset$. 

\section{Adversarial Masked Contrastive Painting}
PainSeg leverages adversarial masked contrastive painting (AMCP) to gradually refine the initial prompt $P$ to the object mask $M$. The AMCP approach is composed of alternating I-steps and O-steps, as illustrated in Figure \ref{fig:teaser}. During each step, a region of the image is masked out based on the previous iteration's mask, and the masked region is then repainted and compared to the original image to refine the mask prediction. To improve the segmentation's robustness, PainSeg introduces adversarial mask updating, which helps to ensure that the mask accurately reflects the object's boundaries. The I-step is used to shrink the object mask by leveraging background consistency, thereby eliminating false-positive regions. On the other hand, the O-step expands the object mask by utilizing object shape consistency to link up false-negative foreground regions. 
\begin{figure}[t]
    \centering    
    \includegraphics[width=\linewidth]{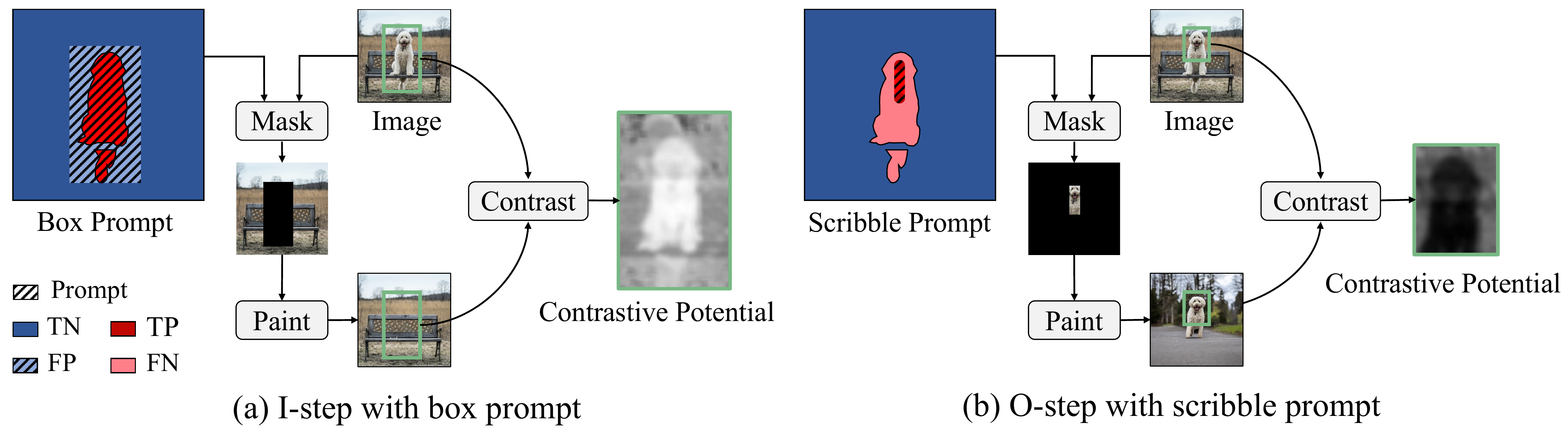}
    \caption{\textbf{Illustration of I-step and O-step with initial prompts.} (a) We show an I-step with a box prompt as the initial mask, where the object region has a significant difference between the original and painted images. (b) We show an O-step with a scribble prompt as the initial mask, where the object region has a small difference between the original and painted images. 
    % Note that we only consider the neighbor near the given prompt when contrasting images.
    }
    % \ca{bg inpaint big diff, out paint small diff, do not mention prompt, later prompt can be formulated as xxx problem}
    \label{fig:pipeline}
\end{figure}

\subsection{Contrastive Painting}
\label{sec:contrastive painting}
% In AMCP, the inpainting \& outpainting mask is complementary.
We first discuss the rationality of segmenting objects by contrasting painted and original images. Given a visual prompt $P$, the relation between the prompted area and object mask can be categorized into three types: background false positive (\cref{fig:pipeline} (a)), foreground false negative (\cref{fig:pipeline} (b)) and a hybrid of both. We tackle the prompt-guided segmentation by separately addressing the background false positive and foreground false negative with I-step and O-step respectively.

We discuss the painted content with different mask situations. To avoid ambiguity, we first denote the generative model taking background and foreground as conditions as inpainting model $\phi(\cdot)$ outpainting model $\psi(\cdot)$ respectively. We consider the prompted area as the initial mask $M_0=P$. When the initial mask has false positives, \ie, $M\subset M_0$, as shown in \cref{fig:pipeline} (a), the inpainted content tends to complete the background based on the background consistency. In this way, the inpainted pixels inside the object will have a significant difference compared to the original image. In contrast, when the initial mask has false negatives, \ie, $M_0\subset M$, as shown in \cref{fig:pipeline} (b), the outpainted content tends to complete the partial object leading to a low difference with the original image inside object region. We notice that I-step can address background false-positive and O-step can address foreground false-negative. By alternating conducting I-step and O-step, we can leverage both foreground and background consistency and address more complicated cases.

\subsection{Contrastive Potential}
\label{sec:contrastive potential}
Given the a image $I$ and a mask $M_t$, we define a contrastive potential $\Phi$ to measure the region relations, which contains three terms 
\begin{equation}
\Phi=\lambda_{paint}\Phi_{paint}+\lambda_{color}\Phi_{color}+\lambda_{prompt}\Phi_{prompt}.
\end{equation}
We introduce a box region $B$ that encloses the foreground and define the $\Phi_{paint}$ term as the distance between the painted and the original image. Specifically, $\Phi_{paint} = B\circ|\mathcal{E}(I)-\mathcal{E}(I_{paint})|_2$, where $\mathcal{E}: \mathbb{R}^{3\times H\times W}\rightarrow\mathbb{R}^{C\times H\times W}$ is a function that projects the image to a high-dimensional space. $\circ$ denotes the Hadamard product.

The $\Phi_{color}$ term measures the pixel-level color similarity inside and outside the mask $M_t$. To compute it, we use the output of the conditional random field algorithm \cite{krahenbuhl2011efficient} and define $\Phi_{color} = \mathcal{C}(M_t\circ B,I\circ B)$, where $\mathcal{C}$ is a function that takes a mask and an image as inputs and outputs the probability of whether a pixel should belong to the masked region.
% Let us denote a bounding box that warps the foreground region as $B$. 
% The $\Phi_{paint}$ is defined as the distance between the in/outpainted area and the original image $\Phi_{paint}=B\circ\|\mathcal{E}(I)-\mathcal{E}(I_{paint})\|_2$ where $\mathcal{E}: \mathbb{R}^{3\times H\times W}\rightarrow\mathbb{R}^{C\times H\times W}$ is a function to project the image to a high-dimensional space. 
% $\Phi_{color}$ is a function to measure the pixel-level color similarity inside and outside the mask $M$.
% Here we leverage the output of conditional random field \cite{krahenbuhl2011efficient} as our $\Phi_{color}=\mathcal{C}(M\circ B,I\circ B)$. $\mathcal{C}$ takes a mask and an image as inputs and outputs the probability of whether a pixel should belong to the masked region. 
% $\Phi_{prompt}$ is a constant controlled by the position of visual prompts to better locate the target object. 
% \paragraph{Prompt priors.}
% For box, scribble, and point prompts, we introduce prompt prior $\Phi_{prompt}$ to the contrastive potential. Let us denote $[x_l,y_l]$ as the coordinates of $l$-th point in the prompt area. The prompt prior is given by
% \begin{equation}
%     \Phi_{prompt}[i,j] = \max_l\,\mathcal{G}(x_l,y_l)[i,j]
% \end{equation}
% where $\mathcal{G}$ is a Gaussian function and $\mathcal{G}[i,j]=\mathrm{exp}(\frac{(i-x_l)^2}{\sigma_x^2}+\frac{(j-y_l)^2}{\sigma_y^2})$. Specifically, for the box prompt, we only take the center point of the box into account.

To further incorporate prompt information, we introduce prompt priors $\Phi_{prompt}$ to the contrastive potential for box, scribble, and point prompts. Let us denote $[x_l,y_l]$ as the coordinates of the $l$-th point in the prompt area. The prompt prior is defined as follows:
\begin{equation}
\Phi_{prompt}[i,j] = \max_l,\mathcal{G}(x_l,y_l)[i,j],
\end{equation}
where $\mathcal{G}$ is a two-dimensional Gaussian function, and $\mathcal{G}[i,j]=\mathrm{exp}(\frac{(i-x_l)^2}{\sigma_x^2}+\frac{(j-y_l)^2}{\sigma_y^2})$. Specifically, for the box prompt, we only take the center point of the box into account. The prompt priors are designed to leverage the positional information of the prompts to better locate the target object. By taking the maximum value of the Gaussian function over all points in the prompt area, $\Phi_{prompt}$ captures the overall strength of the prompt signal.

\subsection{Adversarial I/O-Step}
I-step and O-step share the same mask, paint, and contrast processes while the input mask in I-step is the background mask and, in O-step, the foreground mask. 
% （1）share/diff （2）1 equ to represent，（3）why diff？converge to gt？
Given the original image $I$ and an input mask $M_t$ from the $t$-th step of AMCP (assuming $M_t$ is a background mask thus $t+1$-th step is an I-step), we first filter out the masked region by $I\circ M_t$ and then paint the image $I_{paint}=\phi(I\circ M_t)$. 
% if $t+1$-th step is I-step else $I_{paint}=\psi(I\circ M_t)$
As discussed in \cref{sec:contrastive painting}, the foreground region will have a significant difference between painted and original images.
We obtain the updated mask $M_{t+1}$ by k-means clustering $\mathcal{K}(\cdot)$ over the contrastive potential $\Phi$. Let us denote $\mu_k$ and $S_k\in\{0,1\}^{H\times W}$ as the $k$-th cluster center and its corresponding identity map ($S_k[i,j]=1$ if pixel $[i,j]$ belongs to center $k$ else 0). The updated mask can be found by
\begin{equation}
\begin{aligned}
    M_{t+1}=S_{k^*},\,k^*=\arg\max_k\mu_k.
\label{equ:paint}
\end{aligned}
\end{equation}
% which segments the foreground region in the I-step and the background region in the O-step. 
The updated mask $M_{t+1}$ is a foreground mask thus the next step will be an O-step. Similarly, we paint the image by $I_{paint}=\psi(I\circ M_{t+1})$. Here, the difference in background area will have a significant difference between painted and original images. Thereby, the updated mask $M_{t+2}$ from O-step can be computed using the same rule as \cref{equ:paint} which leads to a background mask.
By updating the mask by \cref{equ:paint}, we notice that when the input mask $M_t$ is a background mask, then the output mask will be a foreground mask and vice versa. Thereby, the alternating I-step and O-step can be automatically achieved. 

% \paragraph{Adversarial constraint.} As discussed in \cref{sec:contrastive painting}, I-step works well for background false-positives and O-step works well for foreground false-negatives. Specifically, we constrain the updated mask to only cut off pixels in I-step. And constrain the updated mask to only link up pixels in the O-step.
As discussed in \cref{sec:contrastive painting}, I-step is advantageous for reducing false positives in the background, whereas O-step is beneficial for reducing false negatives in the foreground. Specifically, the updated mask is configured to only cut off pixels in the I-step, and to only link up pixels in the O-step. Let $M_t^+$ and $M_t^{-}$ as the dilated and eroded masks of $M_t$. We constrain to only update the regions near the foreground-background boundary. In this way, the updating rule for AMCP can be rewritten as
\begin{equation}
    M_{t+1}=\left\{
\begin{aligned}
& S_{k^*}\circ\Delta^{-}+\bar{M_{t}}\circ(1-\Delta^{-}),\quad \text{I-step}\\
& S_{k^*}\circ\Delta^{+}+\bar{M_{t}}\circ(1-\Delta^{+}),\quad \text{O-step}\\
\end{aligned}
\right.
,\,k^*=\arg\max_k\mu_k.
\end{equation}
where $\Delta^{-}=M_t-M_t^{-}$ and $\Delta^{+}=M_t^+-M_t$ are the inner and outer neighbors of $M_t$. 
% By adversarially alternating I-step and O-step, AMCP can address more complicated cases with both false positives and false negatives. In particular, since the generative painting has randomness, we paint the image $N$ times in each step and leverage the averaged mask as output for each step. 
Through the adversarial alternation of I-steps and O-steps, AMCP can handle more complex cases involving both false positives and false negatives. Due to the randomness inherent in generative painting, we paint the image $N$ times in each step, and use the averaged mask as an output.

\subsection{Discussion}
\label{sec:discussion}
In this section, we introduce the mathematical formulation of AMCP. Mathematically, an image can be represented as a masked combination of a foreground image $I_F$ and a background image $I_B$
\begin{equation}
    I=I_F\circ M+I_B\circ \bar{M},\,M\in\{0,1\}^{H\times W\times 1}.
\end{equation}
$M$ is a foreground mask. $\bar{M}=1-M$. 
An inpainting model $\phi[\cdot]$ is defined to generate pixels inside the mask given the pixels outside the mask as a condition.  
Similarly, an outpainting model $\psi[\cdot]$ predicts pixels outside the mask given the pixels inside the mask as a condition. 
In our method, we aim to find a $M$ that maximizes
\begin{equation}
    % \arg\max_M\|M\circ I-\varphi_{in}[(1-M)\circ I]\|-\mathbb{E}_\Delta\|\Delta\circ I-\varphi_{out}[(M-\Delta)\circ I]\|
    \argmax_M\underbrace{\left\|I\circ \Delta^{-}-\phi(I\circ \bar{M})\circ \Delta^{-}\right\|_d}_{\text{\normalsize{I-Step}}}+\underbrace{\left\|I\circ\Delta^+-\psi[I\circ M]\circ\Delta^+\right\|_d}_{\text{\normalsize{O-Step}}}
\label{equ:problem}
\end{equation}
% where $\|\cdot\|_d$ is a distance function. $\Delta^-$ and $\Delta^+$ are the neighbors inside and outside mask $M$ respectively. 
The first term aims to maximize the difference between the original image $I$ and the inpainted image $\phi(I\circ \bar{M})$ in the inner neighbor $\Delta^-$ which corresponds to the I-step in AMCP. The second term aims to maximize the difference between the original image $I$ and the outpainted image $\psi[I\circ M]$ in the outer neighbor $\Delta^{+}$ corresponding to the O-step. 

In each step, our mask, paint, and contrast operations can be considered as an expectation-maximization-like (EM-like) process with the latent variable of $I_{paint}$ to maximize \cref{equ:problem}. On one hand, the $I_{paint}$ is estimated by the mask and paint operations where the conditional probability $p(I_{paint}|I,M)$ is characterized by the generative painting models (expectation step). On the other hand, the predicted mask $M$ can be updated by maximizing the contrastive potential $\Phi$ (maximization step). Since the EM algorithm is sensitive to the initial value, solely updating with I-step or O-step cannot achieve robust performance. With the alternating I-step and O-step, we introduce an adversarial updating process which leads to a more robust mask estimation.

\tabmaskprompt{t}

\section{Experiment}
\subsection{Datasets}
For mask-prompt segmentation, we evaluate on DUTS-TE \cite{wang2017learning} and ECSSD \cite{shi2015hierarchical}. DUTS-TE contains 5,019 images selected from the SUN dataset \cite{xiao2010sun} and ImageNet test set \cite{deng2009imagenet}. ECSSD \cite{shi2015hierarchical} contains 1,000 images that were selected to represent complex scenes. For box-prompt segmentation, we evaluate on PASCAL VOC \cite{everingham2010pascal} val set and COCO \cite{lin2014microsoft} MVAL datasets. COCO MVal contains 800 object instances from the validation set with 10 images from each of the 80 categories.
For point-prompt segmentation, we use three datasets including GrabCut \cite{rother2004grabcut} which contains 50 images and corresponding segmentation masks that delineate a foreground object; Berkeley \cite{mcguinness2010comparative} which contains 96 images with 100 instances with more difficulty than GrabCut and DAVIS \cite{perazzi2016benchmark} which is a video dataset and 10\% of the annotated frames are randomly selected, yielding 345 images that are used in the evaluation
\subsection{Experimental Setup}
\paragraph{Evaluation metrics.}
% To evaluate the segmentation quality, we leverage the intersection over union (IoU) as our metric.
In accordance with previous methods \cite{kirillov2023segment,wang2022tokencut}, we evaluate segmentation quality using intersection over union (IoU).
% Since the coarse mask-prompt segmentation is evaluated on salient object benchmarks, we additionally report pixel accuracy and maximal $F_{\beta}$ score (max $F_\beta$) with $\beta^2$ set to 0.3 following \cite{wang2022self}. 

\paragraph{Implementation details.}
We leverage the inpainting models trained with latent-diffusion pipeline \cite{rombach2021highresolution} as our $\phi$ and $\psi$. We set the diffusion iterations to 50. We leverage DINO \cite{caron2021emerging} pretrained VIT-S/8 \cite{dosovitskiy2020image} as our $\mathcal{E}$. We use \cite{krahenbuhl2011efficient} as our $\mathcal{C}(\cdot)$ to calculate $\Phi_{color}$. If no specification, for all experiments, the masked contrastive painting starts from the I-step and updates for 5 steps. We set the number of cluster centers to 3 in the first three steps for point, box and scribble prompts otherwise 2. We set $\lambda_{paint}=0.8$, $\lambda_{color}=0.2$ and $\lambda_{prompt}=0.2$ if in I-step and $\lambda_{prompt}=-0.2$ if in O-step. We average N=5 painted images to obtain the updated mask for each step. The $\sigma_x$ and $\sigma_y$ are set to $\frac{1}{10}$ of the width and height of the bounding box of the current stage mask respectively. $\Delta^+$ and $\Delta^{-}$ are the neighbors 32 pixels outside and inside the object boundary. We leverage dilation and erosion to filter out sparse points for each iteration. The kernel size is set to 5. For the mask and box prompts, we set the prompt as the initial mask. For the point and scribble prompts, we set the entire image as the initial masked region. The images are padded to $512\times 512$ to fit the generative inpainting model.
% Except for the object region, we also keep the outer counter of each image as the condition in the O-step, as we find it helps to keep the generated image stable. This will not impact the object clustering as we only take the neighbor of the given mask into account.

\tabpointprompt{t}

\subsection{Main Results}
\paragraph{Coarse mask prompt.}
Since the usage of the ground-truth coarse mask as a prompt is rare, we evaluate PaintSeg on two unsupervised salient object detection benchmarks and leverage the coarse mask generated from TokenCut \cite{wang2022tokencut} as our prompt. 
As shown in \cref{tab:saliency-detection}, PaintSeg achieves encouraging performance that is even comparable with training-based methods. Under the training-free setting, PaintSeg significantly outperforms previous methods by a margin of 4.6 IoU on DUTS-TE and 3.4 IoU on ECSSD. We attribute the performance improvement to the error correction capability of PaintSeg. With alternating between I-step and O-step, the proposed PanintSeg can handle noisy prompts effectively. The robustness of PaintSeg will be discussed in more detail in \cref{sec:ablation}.  
%The performance improvement attributes to the error correction capability of PaintSeg. With the alternating I-step and O-step, the proposed PanintSeg can handle noisy prompts. We will discuss more the robustness of PaintSeg in \cref{sec:ablation}. 
% It is also worth mentioning that the proposed PaintSeg achieves inspiring performance that is even comparable with training-based methods. 
% We attribute the good performance to the masked contrastive painting which enables PaintSeg the capability of handling noisy input prompt. Thereby, even the mask prompt from TokenCut is not accurate, our method can correct the 

\paragraph{Point prompt.}
As shown in \cref{tab:point prompt}, we compare our method with state-of-the-art point prompt segmentation approaches. PaintSeg consistently outperforms the training-free methods. Even compared to training-based methods with ground truth supervision, PaintSeg still achieves the best performance on GrabCut and DAVIS datasets. MIS \cite{li2023multi} is an unsupervised approach equipped with second-stage training. We notice that our method can significantly outperform it in terms of IoU, with improvements of 8.2, 6.8, and 16.1 on GrabCut, Berkeley, and DAVIS correspondingly. 
% \CC{This claim is not grounded. There is no 10 IoU difference in each dataset. Please modify.}

\tabboxprompt{t}

\paragraph{Box prompt.}
Since there is no unsupervised box-prompted segmentation that can be directly compared, we compare the proposed method with several baselines including TokenCut \cite{wang2022tokencut}, CutLER \cite{wang2023cut} and MaskRCNN \cite{he2017mask}. We first cut off the ground truth box region and then run the baselines. 
% As shown in \cref{tab:box prompt}, when comparing with training-based Mask-RCNN and CutLER, PaintSeg shows a slightly inferior performance which can be accounted for the lack of training to handle complex scenarios. 
As shown in \cref{tab:box prompt}, when compared with training-based Mask-RCNN and CutLER, PaintSeg shows suboptimal performance, which can be explained by the lack of training to handle complex scenarios. However, as MaskRCNN is trained on 80 COCO object categories, the "unseen" gap remains substantial. PaintSeg provides an alternative solution that is not reliant on training, thus making it more general and capable of handling new categories of objects. When compared with unsupervised approaches, our method eclipses TokenCut by a large margin on both PASCAL VOC and COCO MVal datasets. 
% \XL{do you have any good idea to explain the inferior? My brain is almost blacked out now.. I am thinking explain it from the perspective of not training \& general}
% \CC{This claim is not well-supported. The reviewers won't feel our results are comparable. Please modify.} 

\begin{figure}[t]
    \centering    
    \includegraphics[width=\linewidth]{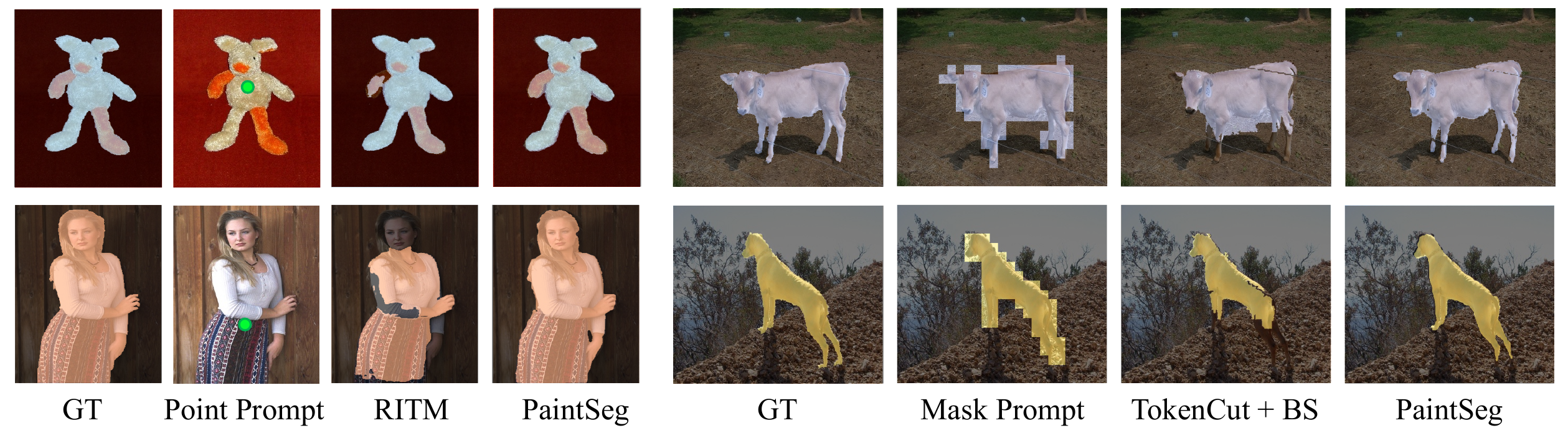}
    \caption{\textbf{Qualitative results of baselines and our PaintSeg with point and mask prompts.} Green point denotes the point prompt. The mask prompt is generated by unsupervised TokenCut \cite{wang2022tokencut}. BS represents the bilateral solver \cite{bilateralsolverbarron2016}.
    We compare with RITM \cite{sofiiuk2022reviving} and TokenCut \cite{wang2022tokencut}.
    }
    \label{fig:vis comparison}
\end{figure}

\paragraph{Qualitative results.}
% We visualize the qualitative results with point and coarse mask prompt in \cref{fig:vis comparison}. PaintSeg successfully segments the complete object while baselines miss parts of the object.
We visualize the qualitative results with point and coarse mask prompt in \cref{fig:vis comparison}. Our visualization depicts comparably reliable results. Comparatively, PaintSeg segments a relatively complete object, while baselines miss some parts of it.
% Notably, even with an inaccurate prompt, \eg, butterfly, PaintSeg can still complete the object mask by AMCP. 

\tabeffectiverobust{t}
\tabdesignchoice{t}

\subsection{Analyses}
\label{sec:ablation}

\paragraph{Module effectiveness in AMCP.}
We step by step add proposed modules in AMCP to validate the effectiveness. As shown in \cref{tab:effectiveness}, we report the results on ECSSD with coarse-mask prompts. We observe that the missing of either step impacts the performance, as evidenced by the significant drop in IoU (compared to alternating I-step and O-step). With the adversarial mask updating constraint, AMCP achieves the best performance of 80.6 IoU.

\paragraph{Robustness of AMCP with different prompts.}
In \cref{tab:prompt sensitivity}, we add noise to the initial prompt by randomly shifting the position to investigate the robustness of AMCP. The scale of random noise is determined, \wrt, half the length of the diagonal of the ground-truth bounding box. We observe that AMCP remains robust and only shows a slight performance drop with a noise rate of less than 30\%. The robust capability can be attributed to 1) the alternating I-step and O-step to leverage both background and object shape consistency, and 2) the adversarial mask updating to tackle the background false-positives and foreground false-negatives.

% \paragraph{Potential curve.}
% As discussed in \cref{sec:discussion}, our method is actually finding a solution maximizing \cref{equ:problem}. We visualize the  normalized contrastive potential curve over each iteration as shown in \cref{fig:potential curve}. We observe that the potential curve increases after each update.

\paragraph{Design choices in AMCP.}
We conduct experiments to ablate the design choices in AMCP and their impacts on the segmentation performance. We first study the effect of cluster center numbers for quantizing contrastive potential. With a larger cluster center, AMCP will ignore more ambiguous regions. As shown in \cref{tab:cluster center}, we notice a cluster center of 2 achieves the best performance for mask prompt. After that, we ablate on the AMCP step number in \cref{tab:step number}. The segmentation performance keeps increasing until reaching a step number of 5. In this way, we choose 5 as our step number. As we leverage the diffusion-based generative model, we ablate the iterations for the diffusion process as it can impact the image quality. As expected, \cref{tab:iter painting} demonstrates that a larger iteration number can reach a better performance. To filter out irrelevant background regions, we crop a box region wrapping the given object mask to contrast images. We ablate the box size in \cref{tab:box size}. We notice that a box slightly larger than the bounding box to the given mask can achieve the best performance. An explanation for this could be that a box tightly enclosing an object will result in a high proportion of object region, which may dominate the features and lead to ambiguity. Properly introducing background can make the extracted features more discriminative and easier for clustering.
%We consider, with a box just fixing the object, the object region will be too large thus dominating the features and leading to ambiguity. Properly introducing background can make the extracted feature more discriminative and easier for clustering.

\begin{figure}[t]
    \centering    
    \includegraphics[width=\linewidth]
    {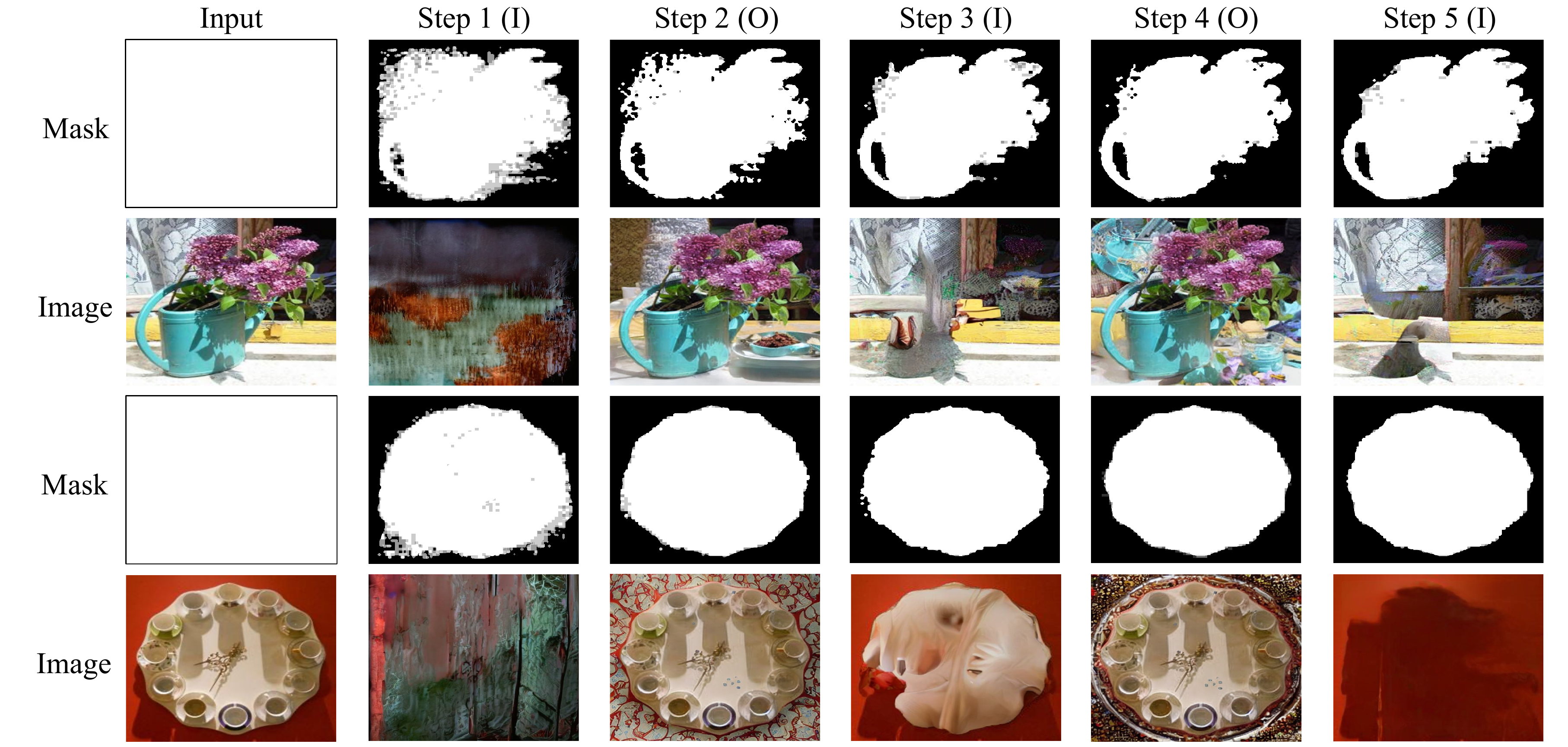}
    \caption{\textbf{Iterative process of AMCP with box prompt.} We inverse the outputted background mask in I-step for better comparison. We only visualize the box prompted area.}
    %\CC{@Xiang. The boundary of the foreground in Step2(O) seems better than Step4(O) in the top example. Also, do we have any better examples to replace the bottom example? There is no need to change if it takes too long.}}
    % \vspace{-5pt}
    \label{fig:vis AMCP}
\end{figure}

% \paragraph{Behavior without prompt.}
% We discuss the behavior of PaintSeg without any visual prompt. Here, we set the initial mask as the entire image and eliminate the prompt prior in calculating contrastive potential.  

\paragraph{Visualization of mask updating.}
To better illustrate the iterative process of AMCP, as shown in \cref{fig:vis AMCP}, we visualize the averaged mask output (among $N$ painted images in each step) for each step with a box prompt. As the given mask only contains background false positives, I-step plays a major role to cut false-positive backgrounds in AMCP. The mask shrink can also be observed after the O-step which is due to the binarization of the averaged mask from the I-step instead of contrastive painting. We observe that the updated masks are gradually closer to the mask of the target object with AMCP.
%We observe that the initial mask gradually moves to the mask of the target object with AMCP.
% The lower case suffers from an error during the first I-step, we can observe that the error is gradually corrected in the following steps.

% \section{Limitations and Boarder Impact}
% \paragraph{Limitations.} The proposed PaintSeg highly relies on the object information provided by visual prompts and is lack of the capability of detection in a complex scene. In fact,  
\section{Conclusion}
% In conclusion, the paper presents PaintSeg, an unsupervised and training-free approach for image object segmentation that leverages off-the-shelf generative models. The proposed approach utilizes an adversarial masked contrastive painting (AMCP) process, which creates a contrast between the original image and a painted image by alternating between I-step (inpainting) and O-step (outpainting). The I-step and O-step leverage the background and object shape consistency to gradually advance the object mask closer to the ground truth. PaintSeg can deal with inaccurate initial masks and adapt to various visual prompts, such as coarse masks, bounding boxes, scribbles, and points. Extensive experiments were conducted on seven different image segmentation datasets, which demonstrated the superiority and generalization ability. Overall, PaintSeg bridges the gap between generative models and segmentation and provides a robust and training-free approach for unsupervised image object segmentation. 
To conclude, PaintSeg bridges the gap between generative models and segmentation. It is designed to provide a robust and training-free approach to unsupervised image object segmentation. With the proposed adversarial masked contrastive painting (AMCP) process, PaintSeg creates a contrast between the original image and the painted image by alternately applying I-steps (inpainting) and O-steps (outpainting). The alternating I-step and O-step gradually improve the accuracy of the object mask by leveraging consistency in the background and the shape of the object. The competitiveness of our method on seven different image segmentation datasets suggests that PaintSeg can deal with inaccurate initial masks and adapt to various visual prompts, such as coarse masks, bounding boxes, scribbles, and points. An extensive ablation analysis indicates a number of key factors and advantages of the proposed model, including its design choices and generalizability.
\paragraph{Limitation.}
% PaintSeg has achieved high performance for training-free image segmentation with heterogeneous visual prompts while it currently lacks object discovery capability and thus cannot automatically segment instance-level masks in an image. The discovery capability can be developed by conducting a second-stage training on the segmentation results generated by PaintSeg which will be our future research focus. 
In spite of PaintSeg's high performance for training-free image segmentation with heterogeneous visual prompts, it does not possess object discovery capabilities and therefore cannot automatically recognize instance-level masks in an image. Developing discovery capability can be achieved by conducting second-stage training on the segmentation results generated by PaintSeg, which is our future research focus.

\bibliographystyle{plain}
\bibliography{src/ref}

\renewcommand{\thetable}{{\Alph{table}}}
\renewcommand{\thefigure}{{\Alph{figure}}}
\renewcommand{\thesection}{{\Alph{section}}}
\setcounter{figure}{0}
\setcounter{table}{0}
\setcounter{section}{0}
\clearpage
\section{More Comparison with Mask-RCNN}

\begin{figure}[h!]
    \centering    
    \includegraphics[width=\linewidth]{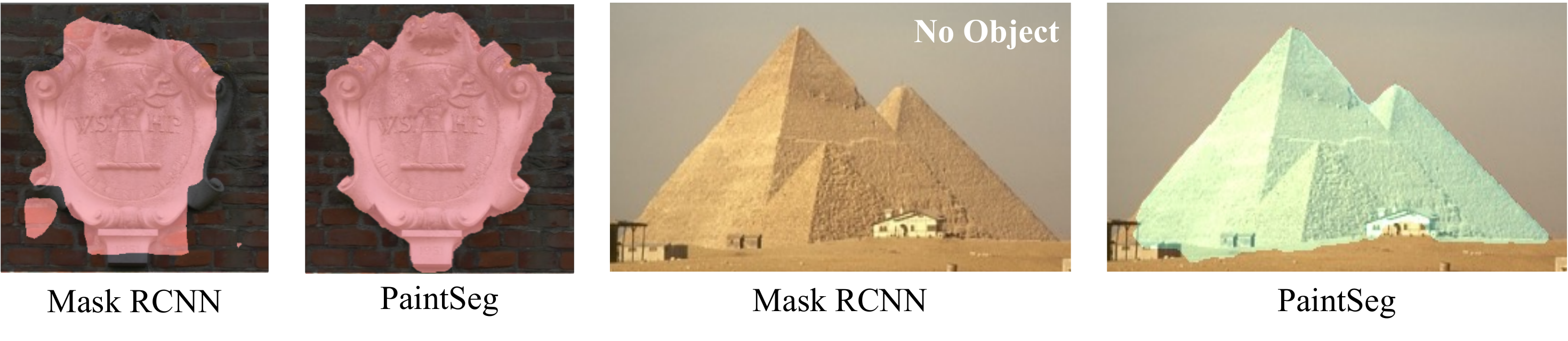}
    \caption{\textbf{Comparison with Mask RCNN with objects beyond 80 COCO categories.}
    }
    \label{fig:maskrcnn}
\end{figure}
We present more results compared with supervised Mask RCNN \cite{he2017mask}. As shown in \cref{fig:maskrcnn}, we compare box-prompted segmentation with Mask RCNN on objects beyond 80 COCO categories. In the shown examples, we observe that Mask RCNN has difficulty segmenting the correct shape of the object. Instead, PaintSeg provides more accurate object segmentation. As Mask RCNN is only trained on 80 COCO object categories, there is still a substantial gap between the seen and the unseen. In contrast, PaintSeg is a solution that does not require training, which makes it more general and capable of handling new object categories.
%We notice Mask RCNN fails to segment the correct object shape. Instead, PaintSeg achieves accurate object segmentation. Since Mask RCNN is only trained on 80 COCO object categories, the ``unseen'' gap remains substantial. PaintSeg, in contrast, provides an alternative solution that is not reliant on training, thus making it more general and capable of handling new categories of objects.

\section{More Ablation Experiments}
In this section, we provide additional ablation studies to illustrate the design choices of PaintSeg.

\begin{table}[h]
\centering
\begin{tabular}{ccccccc}
\toprule
N & 1 & 2 & 3 & 4 & 5 & 6\\
\midrule
IoU & 78.8 & 79.2 & 79.6 & 80.1 & 80.6 & 80.8 \\
\bottomrule
\end{tabular}
\vspace{5pt}
\caption{\textbf{Ablation study on the painted image number $N$ for each step}.}
\label{tab:painted number}
\vspace{-5pt}
\end{table}

\subsection{Sampling Number for Each Step}
We average $N$ painted images in each step to obtain the final mask prediction due to the randomness of the generative painting model. We present an ablation study to illustrate the impact of the number of painted images in each step. As shown in \cref{tab:painted number}, we report the performance on the ECSSD \cite{shi2016ecssd} dataset with coarse mask prompt from TokenCut \cite{wang2022tokencut}. We notice that the performance gradually improved with more painted images averaged in each step. As there is no significant difference in performance between five or six painted images used, we set the number of painted images to five in the PaintSeg process.%We set the number of painted images to five in the PaintSeg process. %We choose a painted image number 5 in PaintSeg.

\begin{table}[h]
\centering
\begin{tabular}{cccccc}
\toprule
DINO \cite{caron2021dino} VIT-S/8 & DINO-V2 \cite{oquab2023dinov2} VIS-S/14\\
\midrule
80.6 & 80.0 \\
\bottomrule
\end{tabular}
\vspace{5pt}
\caption{\textbf{Ablation study on image projector $\mathcal{E}$ used in AMCP}.}
\label{tab:image projector}
\vspace{-5pt}
\end{table}

\subsection{Image Projector}
We conduct an ablation study on image projector $\mathcal{E}$ as illustrated in \cref{tab:image projector}. We compare the widely used DINO \cite{caron2021dino} VIT-S/8 and the latest DINO \cite{oquab2023dinov2} VIS-S/14. The results demonstrate that DINO with a small patch size achieves better performance. It follows that we consider a smaller patch size since PaintSeg requires fine-grained visual information. A larger patch size will blur the object boundary, resulting in a performance drop.%We consider the smaller patch size as the primary reason for the performance superiority of DINO as PaintSeg requires fine-grained visual information. A larger patch size will blur the object boundary thus leading to a performance drop.

\section{More Potential Application}
In this section, we discuss more potential applications of PaintSeg beyond prompt-guided object segmentation.
\begin{figure}[h]
    \centering
    \includegraphics[width=\linewidth]{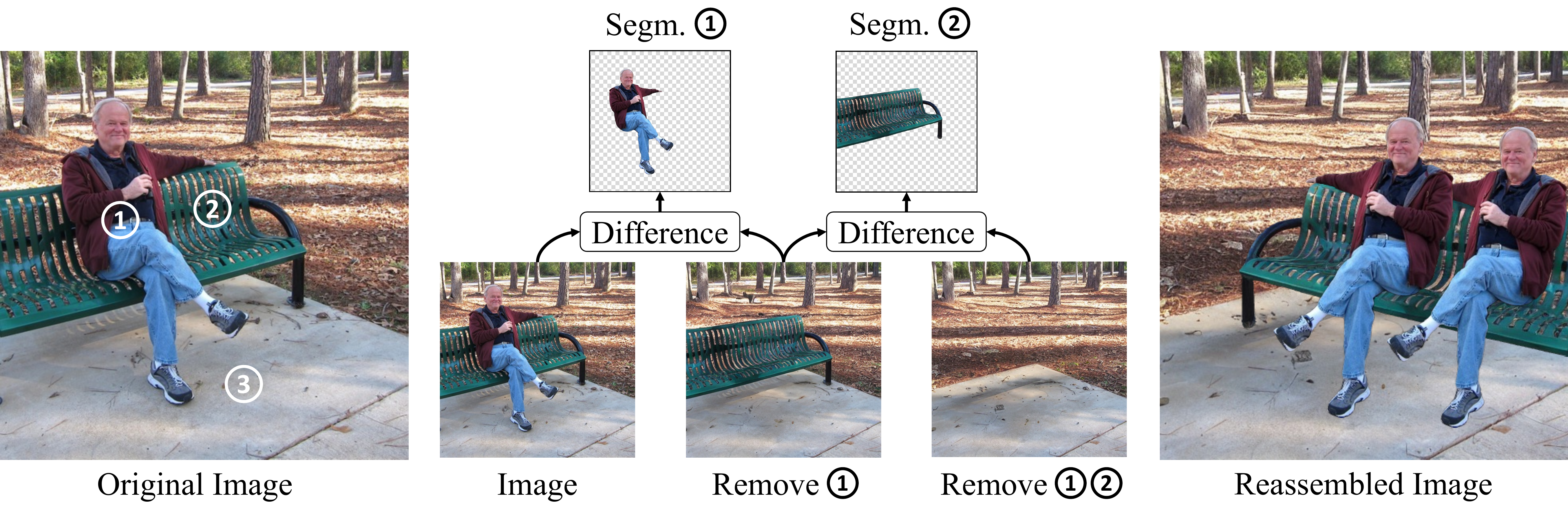}
    \caption{\textbf{Potential application in image edition and amodal segmentation.} PaintSeg can step-by-step remove objects in the image by using the painted image in I-step. With the segmented object and painted image without objects, we can freely assemble them into a new image. Further, PaintSeg supports amodal segmentation, with the painting capability enabling segmentation of the occluded areas.} %Furthermore, thank to the painting capability of PaintSeg, even the occluded area can be segmented enabling PaintSeg to handle amodal segmentation.}
    \label{fig:potentialapp}
\vspace{-5pt}
\end{figure}

\subsection{Image Edition}
In the I-step of AMCP, the painted image will remove the target object while keeping all other contents in the image. In this way, with the segmented objects and an image without target objects, we can reassemble them into a new image as shown in \cref{fig:potentialapp}.

\subsection{Amodal Segmentation}
As shown in \cref{fig:potentialapp}, PaintSeg can layer-by-layer segment objects. By using the painted image in I-step as the input to the next iteration, PaintSeg can attach the amodal capability. We notice that the bench is occluded by the men in \cref{fig:potentialapp}. With the PaintSeg, the full shape of the bench can be segmented.

\section{More Discussion about PaintSeg}
\label{sec:EM}
In PaintSeg, we introduce a latent variable $I_{paint}$ which is characterized by an off-the-shelf generative model $p(I_{paint}|I\circ M)$ conditioned on an image $I$ and a mask $M$. $\circ$ represents Hadamard product. In our method, we leverage the AMCP process to estimate and convert the latent variable $I_{paint}$ into mask prediction $M$ with alternating I-step and O-step. Mathematically, both I-step and O-step can be formulated as an expectation-maximization-like process.

\begin{itemize}
    \item \textbf{Expectation}: We introduce a latent variable $I_{paint}$ in the proposed PaintSeg which is modeled by an off-the-shelf generative painting model $p(I_{paint}|I\circ M)$. We assume the generative model will pick the most likely outcome $I_{paint}$ given $I$ and $M$ for every step.
    \item \textbf{Maximization}: After obtaining the latent variable $I_{paint}$, we define a contrastive potential $\Phi$ and utilize clustering to binarize the mask. Mathematically, the contrasting and clustering processes maximize a posteriori probability $p(M|I_{paint}, I)=e^{-\frac{1}{\|M\|_0}\Phi(I_{paint},I,M)}$.
\end{itemize}

Although we term I-step and O-step separately, they can be formulated as the same EM process. PaintSeg advances the predicted mask to the ground truth by iteratively conducting the EM process in each step.

% \begin{figure}[htp]
%     \centering
%     \includegraphics[width=\linewidth]{figs/noprompt.pdf}
%     \caption{\textbf{Illustration of the behavior of PaintSeg without any prompt.} Given an image and an initial mask including the entire image region as input (without $\Phi_{prompt}$), we notice PaintSeg output a mask includes all foreground objects in the image.  We further run another AMCP process on the masked region (only consider the mask region during clustering) and find that the segmented foreground objects can be further separated. We consider the behavior of PaintSeg demonstrates its potential for automatically analyzing images without prompt input which will be our future research focus.
%     }
%     \label{fig:noprompt}
% \end{figure}

% \section{Behavior without Prompts}
% We discuss the behavior of PaintSeg without any prompt in this subsection. In the initial stage, we let the entire image region as the mask and calculate the contrastive potential without prompt prior. As shown in \cref{fig:noprompt}, we notice that PaintSeg tends to segment all foreground objects exited in the image if no prompt is given. We further run another AMCP process on the masked region (only consider the mask region during clustering) and find that the segmented foreground objects can be further separated. We consider the behavior of PaintSeg demonstrates its potential for automatically analyzing images without prompt input which will be our future research focus.

\section{Difference with Previous Segmentation Approaches}

In this section, we discuss the major differences between the proposed PaintSeg and previous object segmentation methods as follows.

\paragraph{Discriminative \textit{v.s.} Generative\,+\,Discriminative .}
Conventional object segmentation is a discriminative task that leverages a neural network $\theta$ to model the conditional probability of the object mask $M$ given the image $I$ as condition $p_{\theta}(M|I)$. In PaintSeg, we have mask, paint, and contrast operations in each step. Specifically, in paint operation, we enroll a generative model to estimate painted image $I_{paint}$ with mask $M$ and image $I$ as conditions.
After that, the mask can be obtained by comparing the generated image with the original one with a contrastive potential $\Phi$. As discussed in \cref{sec:EM}, the paint operation is a generative process to estimate latent variable $p(I_{paint}|I\circ M)$ and the contrast operation is a discriminative process to obtain a mask prediction based on $p(M|I_{paint},I)$. PaintSeg achieves training-free by constructing a bridge to generative painting models which permits object shape consistency and background content consistency.

\paragraph{Pixel \textit{v.s.} Pixel difference.}
Conventional object segmentation leverages a network to project an image to the feature space and then binarize (cluster) each pixel into foreground or background classes. Differently, instead of directly clustering over the input image, PaintSeg utilizes the difference between the painted and original image, as a proxy, to leverage the object shape prior and background consistency. The contrastive scheme is rooted in the decomposable nature of images and paves a way to incorporate generated images to segment objects.

\paragraph{Training \textit{v.s.} Training-free.}
Conventional object segmentation approaches train the neural network to segment objects requiring time-consuming and expensive data labeling. Some unsupervised segmentation methods \cite{benny2020onegan,chen2019unsupervised,bielski2019emergence,voynov2021biggan} find a segment from a generative model while they typically require training a network on top of the generative model. Instead, our method is a training-free unsupervised method that learns to segment objects from a generative painting model. We consider the PaintSeg provides a way to bridge the generative model and segmentation which may inspire future research.

% \begin{figure}[t]
%     \centering
%     \includegraphics[width=\linewidth]{figs/prompts.pdf}
%     \caption{\textbf{Illustration of error types of all prompts.}
%     }
%     \label{fig:prompts}
% \end{figure}

% \section{Illustration of Prompts}
% We demonstrate the error types suffered from different visual prompts as shown in \cref{fig:prompts}. PaintSeg leverages adversarial masked contrastive painting (AMCP) to robustly address false positives and false negatives thus leading to an accurate segmentation result.

\section{Failure Case Analysis}
\begin{wrapfigure}{r}{0.6\textwidth}
\centering
\vspace{-0.3cm}
\includegraphics[width=0.6\textwidth]{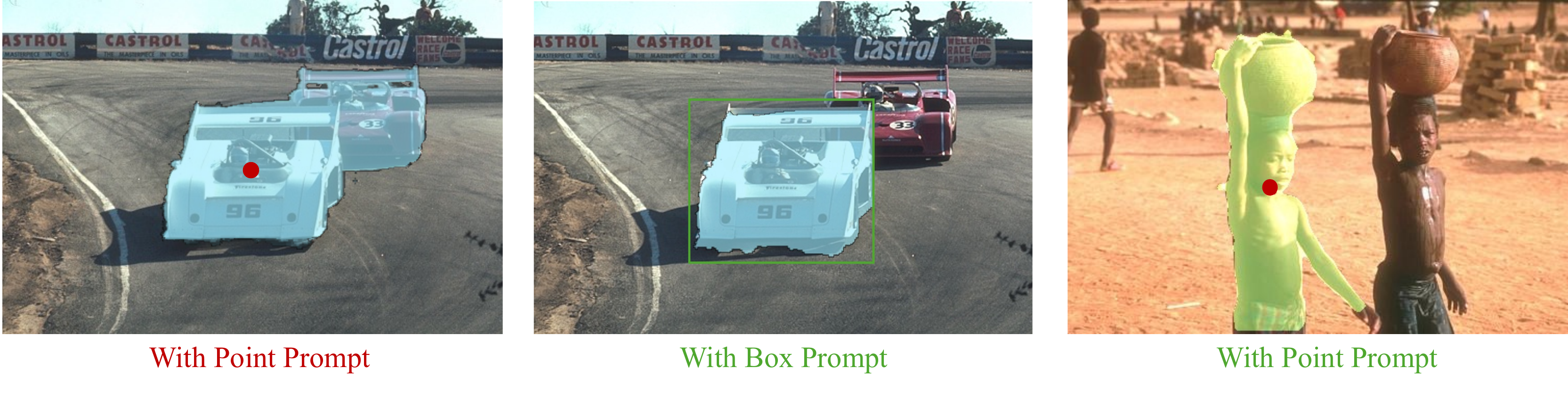}
\caption{Illustration of failure case.}
\vspace{-0.4cm}
\label{fig:failure}
\end{wrapfigure}
We analyze the failure case here. As shown in \cref{fig:failure}, we visualize a failure case when using a point as the prompt. We notice the adjacent car is segmented as a false positive, which is due to the semantic and visual similarity between the target and false positive cars. Despite our method is capable of handling multiple objects with point prompt (right of \cref{fig:failure}), crowded scenarios can make it difficult to segment the accurate object boundary. However, the issue can be overcome through box prompt.
%Although our method can tackle multiple objects in the image with point prompt (right of \cref{fig:failure}), when the scenario goes crowded, our method may fail to segment the accurate object boundary.

\section{More Visualization}
\begin{figure}[t]
    \centering
    \includegraphics[width=\linewidth,height=1.6\linewidth]{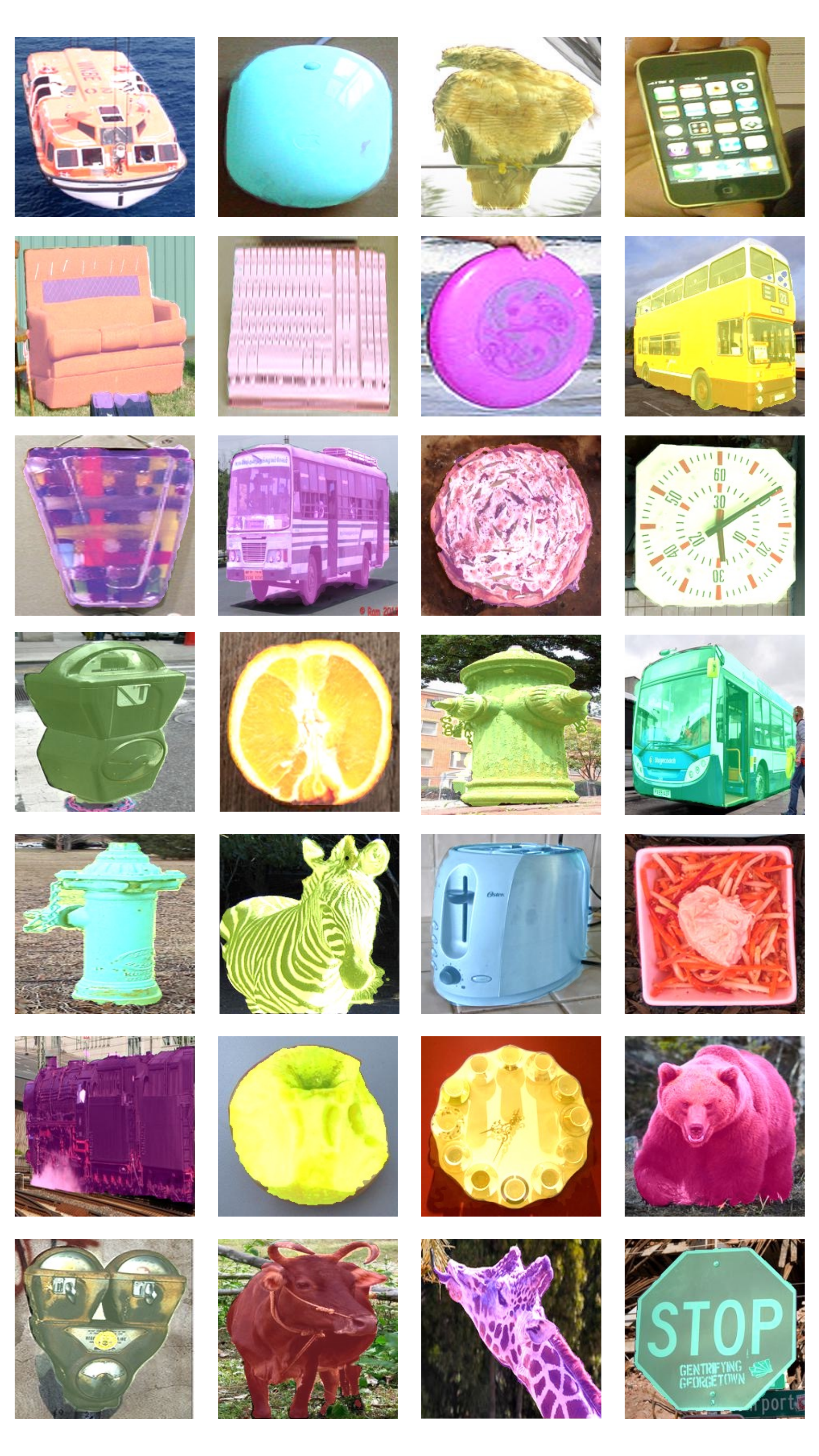}
    \caption{\textbf{More visualization of PaintSeg with box prompt on COCO MVal.}
    }
    \label{fig:box vis}
\end{figure}

\begin{figure}[t]
    \centering
    \includegraphics[width=\linewidth,height=1.6\linewidth]{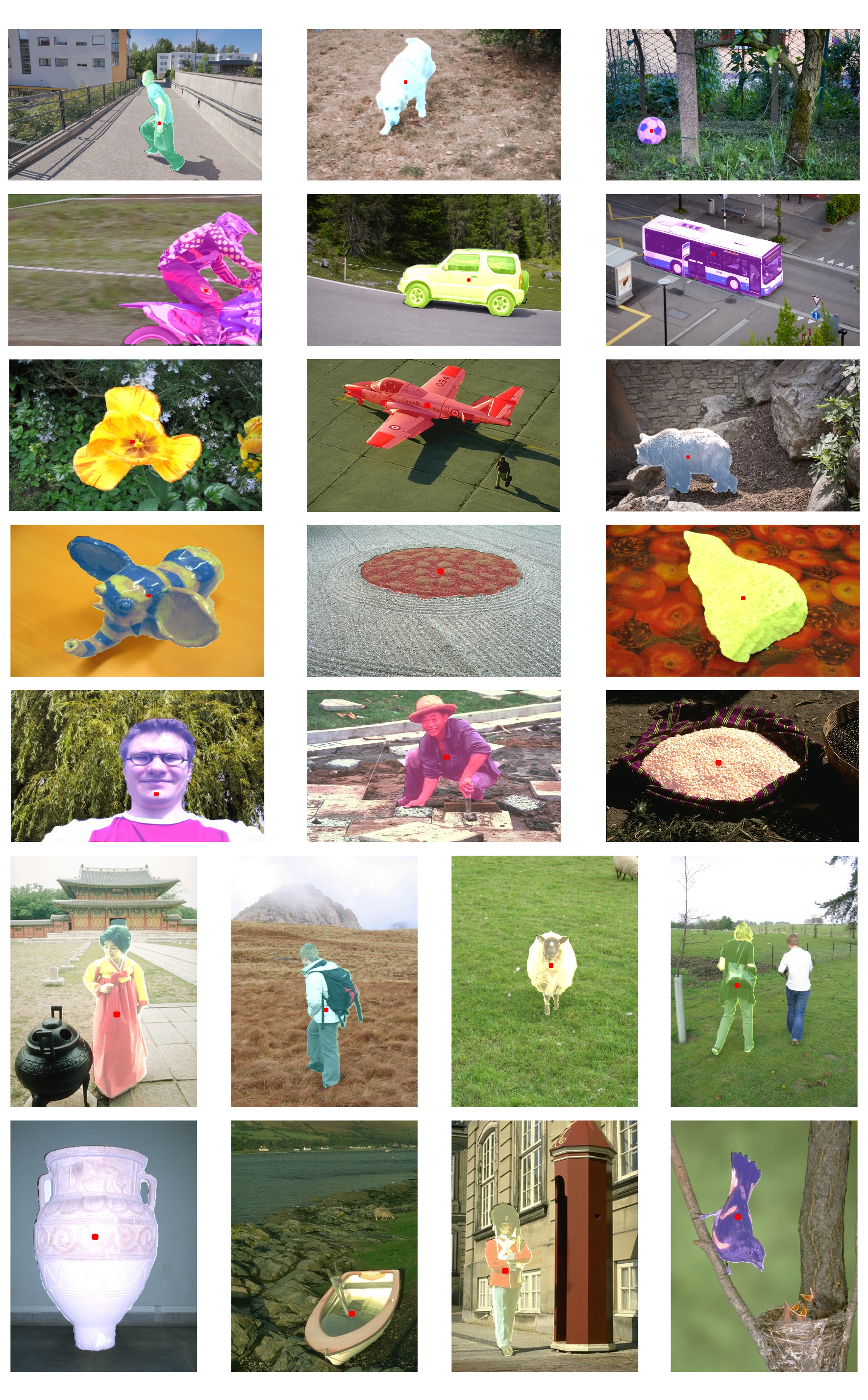}
    \caption{\textbf{More visualization of PaintSeg with point prompt. The point prompt is illustrated by the {\color{red}{red point}} on the image on DAVIS and Berkeley and GrabCut.}
    }
    \label{fig:point vis}
\end{figure}

\begin{figure}[t]
    \centering
    \includegraphics[width=\linewidth,height=1.6\linewidth]{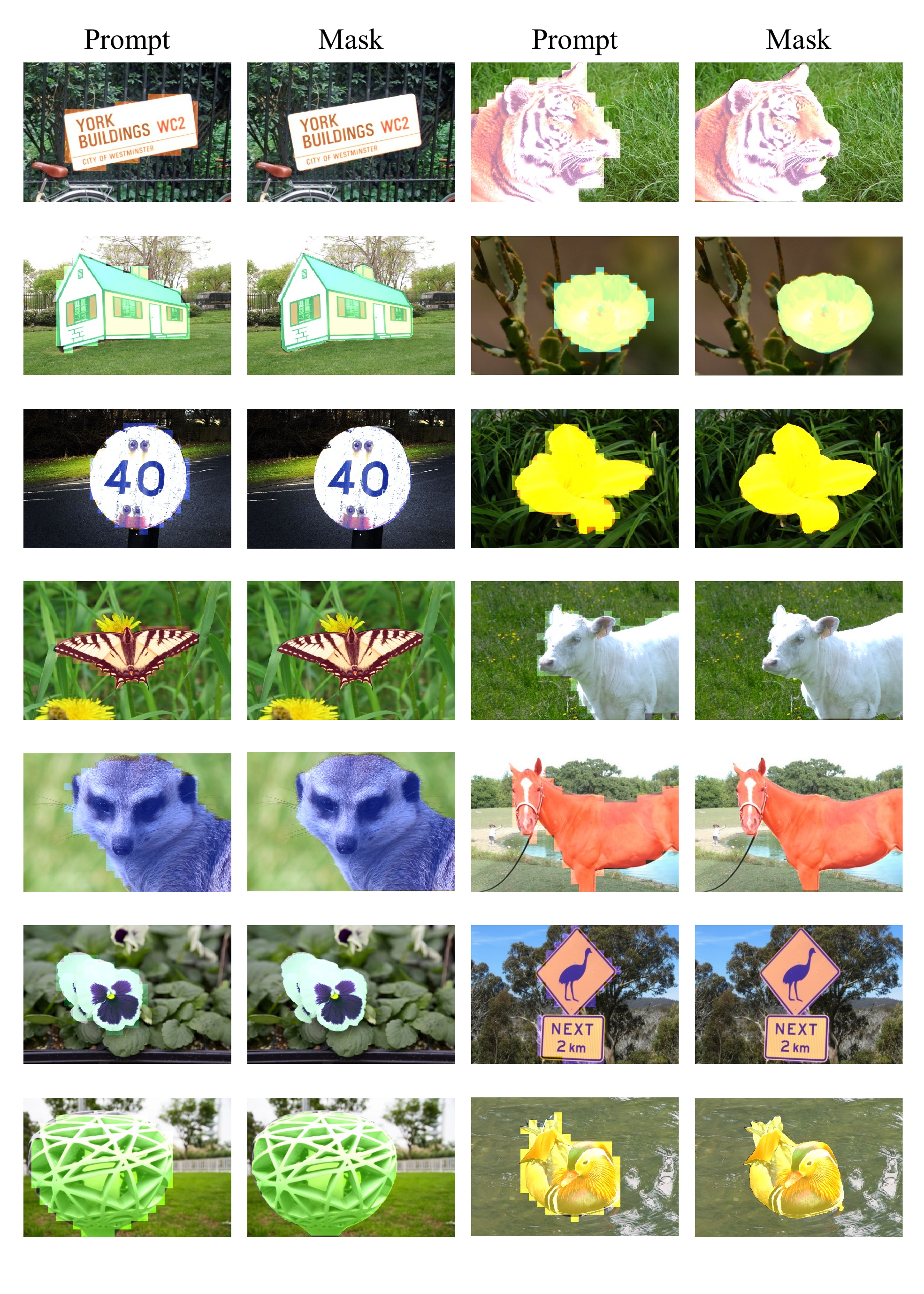}
    \caption{\textbf{More visualization of PaintSeg with coarse mask prompt on ECSSD.}
    }
    \label{fig:mask vis}
\end{figure}

\begin{figure}[t]
    \centering
    \includegraphics[width=\linewidth,height=1.6\linewidth]{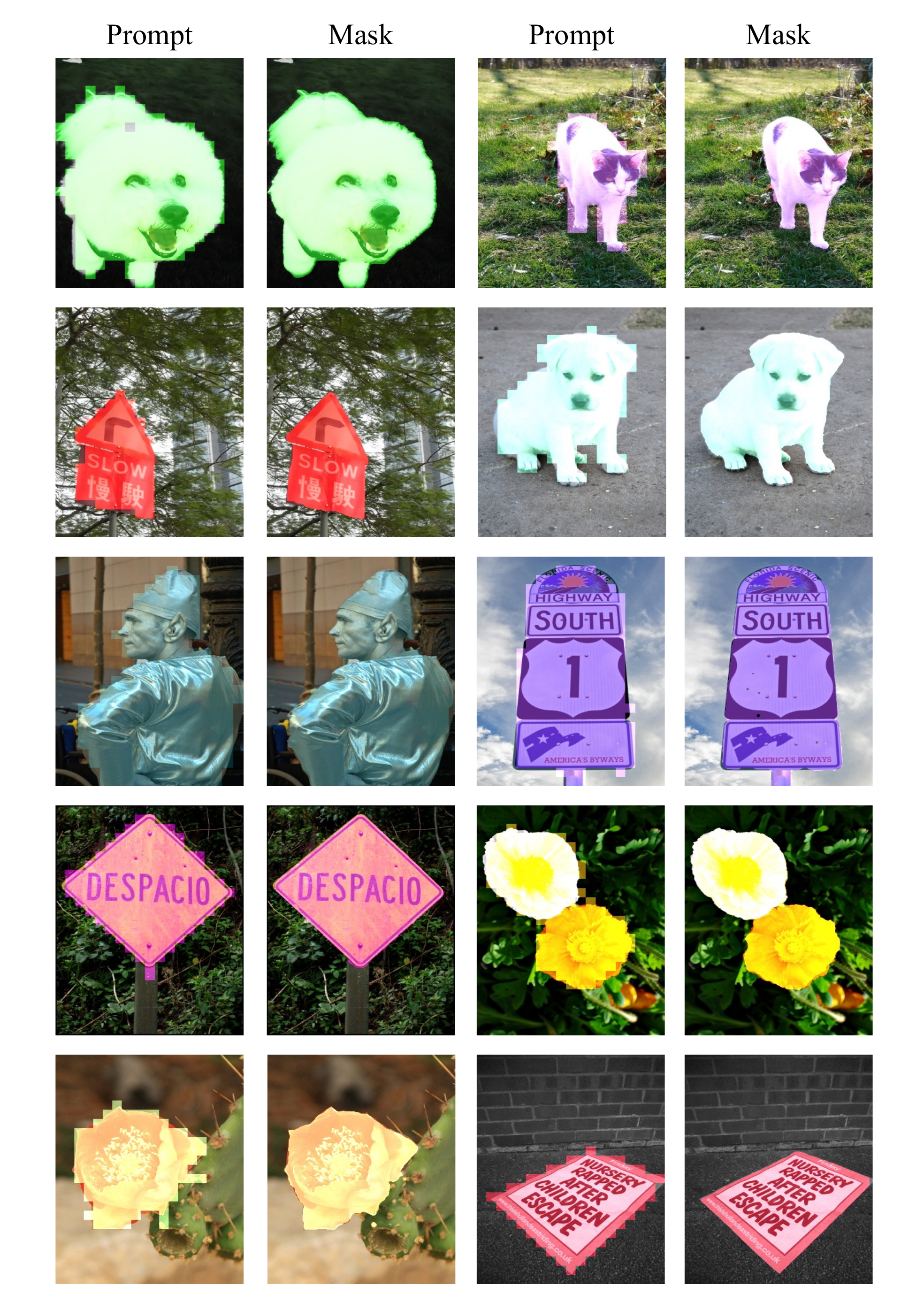}
    \caption{\textbf{More visualization of PaintSeg with coarse mask prompt on ECSSD.}
    }
    \label{fig:mask vis2}
\end{figure}

In this section, we demonstrate more visualization of PaintSeg. We show more qualitative results with box prompt in \cref{fig:box vis}, with point prompt in \cref{fig:point vis} and with coarse mask prompt in \cref{fig:mask vis,fig:mask vis2}.

\end{document}